\documentclass[preprint,12pt]{elsarticle}

\usepackage{geometry}
\usepackage{amssymb}
\usepackage{amsmath}
\usepackage{amsfonts}
\usepackage{subcaption}
\usepackage{bm}
\usepackage{placeins}
\usepackage{graphicx}
\usepackage{lineno}

\newcommand{\refeq}[1]{Eq.~\eqref{#1}}

\newcommand{\mat}[1]{\mathbf{#1}}
\newcommand{\set}[1]{\mathcal{#1}}
\newcommand{\pnorm}[1]{\lVert{#1}\rVert}

\newcommand{\classifier}{\ensuremath{o}}
\newcommand{\setX}{\ensuremath{\set{X}}}
\newcommand{\setY}{\ensuremath{\set{Y}}}
\newcommand{\xorig}{\ensuremath{\vec{x}_{\text{orig}}}}

\newcommand{\y}{\ensuremath{y}}
\newcommand{\ycf}{\ensuremath{y}_{\text{cf}}}
\newcommand{\xcf}{\ensuremath{\vec{x}_{\text{cf}}}}

\newcommand{\RN}{\mathbb{R}}
\newcommand{\dimsym}{d}

\newcommand{\regression}{\ensuremath{f}}

\newcommand{\numSensors}{n_s}
\newcommand{\timeVar}{k}

\DeclareMathOperator*{\regularization}{\ensuremath{{\theta}}}
\DeclareMathOperator*{\loss}{{\ell}}

\newcommand{\refdef}[1]{Definition~\ref{#1}}
\newtheorem{definition}{Definition}

\journal{}

\begin{document}

\begin{frontmatter}



\title{Interpretable Event Diagnosis\\in Water Distribution Networks}

 \author[label1,label2]{Andr\'e~Artelt}
 \author[label2]{Stelios~G.~Vrachimis}
\author[label2]{Demetrios~G.~Eliades}
 \author[label1]{Ulrike~Kuhl}
  \author[label1]{Barbara~Hammer}
 \author[label2,label3]{Marios~M.~Polycarpou}
 
 \affiliation[label1]{organization={Bielefeld University},
             addressline={Inspiration 1},
             city={Bielefeld},
             postcode={33615},
             state={NRW},
             country={Germany}}

\affiliation[label2]{organization={KIOS Research and Innovation Center of Excellence, University of Cyprus},
            addressline={Panepistimiou 1, Aglantzia},
             city={Nicosia},
            postcode={2109},
             country={Cyprus}}

\affiliation[label3]{organization={Electrical and Computer Engineering Department, University of Cyprus},
            addressline={Panepistimiou 1, Aglantzia},
             city={Nicosia},
            postcode={2109},
             country={Cyprus}}

\begin{abstract}
The increasing penetration of information and communication technologies in the design, monitoring, and control of water systems enables the use of algorithms for detecting and identifying unanticipated events (such as leakages or water contamination) using sensor measurements.
However, data-driven methodologies do not always give accurate results and are often not trusted by operators, who may prefer to use their engineering judgment and experience to deal with such events.

In this work, we propose a framework for interpretable event diagnosis --- an approach that assists the operators in associating the results of algorithmic event diagnosis methodologies with their own intuition and experience.
This is achieved by providing contrasting (i.e., counterfactual) explanations of the results provided by fault diagnosis algorithms; their aim is to improve the understanding of the algorithm's inner workings by the operators, thus enabling them to take a more informed decision by combining the results with their personal experiences.
Specifically, we propose \textit{counterfactual event fingerprints}, a representation of the difference between the current event diagnosis and the closest alternative explanation, which can be presented in a graphical way.
The proposed methodology is applied and evaluated on a realistic use case using the L-Town benchmark.
\end{abstract}

\begin{keyword}


Water Distribution Networks \sep Event Diagnosis \sep eXplainable Artificial Intelligence \sep Counterfactual Explanations \sep Interpretability
\end{keyword}

\end{frontmatter}


\section{Introduction}
When an event, such as a leakage, occurs in a Water Distribution Network (WDN), this can affect the dynamics of the system by causing changes in the pressures and flows \cite{Giustolisi2008}. These changes can be monitored by flow and pressure sensors installed within WDNs. Typically, a limited number of flow sensors are installed at the entrance of District Metered Areas (DMAs) to monitor the overall water inflow in the area \cite{Eliades2011a}, while a larger number of pressure sensors (due to reduced capital and installation costs) are installed at certain locations within the DMA to improve leakage detectability \cite{Cuguero-Escofet2017}. 

\subsection{Detecting Events} 
Human operators require automated systems to assist in monitoring for unanticipated events in water systems due to the complexity and large amount of information received by telemetry systems.
Without the use of leakage diagnosis algorithms, detection of leakages in the system may be sometimes performed visually or through the use of some predefined thresholds on sensor measurements based on the operators' experience; yet, in practice, several problems arise: 
1) smaller or slowly developing leakages may be indistinguishable from noise; 
2) the location of the leakage may be far away from sensors and therefore not easily be detectable; 
3) there may be multiple leakages occurring at the same time at different locations; 
4) there may be sensor faults affecting one or more sensors, influencing the decision of whether there is a leakage in the system; 
5) the operator may be biased on how the system should operate, which can affect their decision \cite{Reppa2016}. 

Recent literature on WDN event diagnosis demonstrates the use of automated model-based approaches~\cite{Pecci2020,Vrachimis2021}, or data-driven methodologies \cite{Zou2019,Soldevila2021,peng2024novel,ISLAM2024200404}, which utilize machine learning (ML) methodologies such as deep neural networks \cite{Zhou2019,Abokifa2019}, to assist operators in expanding their detection and diagnosis capabilities.
These automated systems significantly improve leakage detection rates; however, they are not perfect and still require human supervision \cite{Wu2017a}.
Operators who lack insight into the reasons behind an algorithm's classification of events as normal or faulty will tend to perceive these systems as having a higher miss rate, influenced by the inherent principle of loss aversion \cite{Kahneman1991}. A more effective strategy would involve these systems explaining their results to build users' trust \cite{MUSTAFA2024200430}.

\subsection{Interpretable AI}
According to EU's White Paper on AI \cite{whitepaper}, transparency is a major requirement of potentially critical artificial intelligence (AI) based decision-making systems that are deployed in the real world. Transparency in AI refers to the ability to understand (to a certain degree) how a decision was derived \cite{transparency}.
The necessity of this was also recognized by policymakers and the requirement of transparency found its way into legal regulations such as the EU's GDPR~\cite{GDPR} that grants ``a right to an explanation'' to the users. 
Furthermore, the European AI-Act~\cite{euAiAct21} classifies the application of AI methods to critical infrastructures, such as water distribution networks, as high-risk systems; this means that AI-based decision-making systems deployed in these infrastructures must fulfill special requirements concerning security and reliability (e.g. safety bounds for operation, an accountability framework that details human oversight, etc.).
Interpretability refers to ``the degree to which a human can understand the cause of a decision''~\cite{miller2019explanation} and therefore implements a particular way of transparency. In the AI and machine learning (ML) community, the terms ``interpretability'' and ``transparency'' are often used as related terms or interchangeably.

Explanations are a common approach for realizing transparency of an AI system; i.e., achieving transparency by providing explanations in relation to the system's behavior to the user.
Currently, there is significant research on explanations for AI shaping the field of eXplainable AI (XAI)~\cite{ExplainingBlackboxModelsSurvey,molnar2019}. Although a lot of different explanation methodologies have been developed, there is no universal agreement on what constitutes a good explanation~\cite{offert2017i}. Indeed, no single explanation method can fulfill the specific requirements in every situation; instead, one has to carefully select appropriate explanation methods for each use case separately
\cite{DBLP:journals/tamd/RohlfingCSMBBEG21}.
This is also mirrored by the huge variety of XAI technologies that have been proposed in the literature.

At a high level, existing explanation methods can be grouped into \textit{post-hoc explanations} and \textit{self-explanatory systems}~\cite{molnar2019}. Post-hoc explanations are created after the systems' behavior has been observed, while self-explanatory systems are transparent by design. Note that some methods that are considered self-explanatory (e.g.\ decision trees, linear models, etc.), might face challenges, e.g., if their complexity gets too high; hence those may not be ``truly'' self-explanatory for different user groups and scenarios in practice. Post-hoc methods can be further divided into general model-agnostic methods (i.e., methods applicable to any possible AI system) and model-specific methods (i.e., methods tailored towards a specific class of models). A further possible categorization is in between global and local explanations. Global explanations aim to explain the entire AI system, while local explanations focus on specific regions or cases only. 
Prominent instances of post-hoc explanations are feature-based explanations such as LIME~\cite{Lime} (local, model agnostic), LRP~\cite{bach2015pixel} (local, model specific to DNNs), or other 
feature importance values~\cite{FeatureImportance}.
Another prominent instance of post-hoc explanations are example-based methods such as global explanations by prototypes \& criticisms~\cite{PrototypesCriticism} or local explanations such as contrasting explanations and counterfactual explanations~\cite{CounterfactualWachter,CounterfactualReviewChallenges}.

In the following, we focus on contrasting explanations and counterfactual explanations as an intuitive and flexible XAI scheme. 
We will focus on the challenge of how to adapt this general scheme to the specific challenge of 
informed decision-making for fault and leakage
detection in water distribution systems.

\subsection{Contributions}
This work introduces the concept of \textit{interpretable event diagnosis} for water distribution systems, an innovative framework for providing explanations to water systems operators to increase the understanding and trustworthiness of the results provided by event diagnosis algorithms.
To our knowledge, this is the first work aimed at developing methods for interpretable event diagnosis in drinking water distribution systems.
Our key contribution is the development of an explanation methodology called \textit{counterfactual event fingerprints} for designing interpretable event diagnosis algorithms for water distribution networks.

The remainder of this work is structured as follows: First, we review the foundations of counterfactual explanations (Section~\ref{sec:foundations:counterfactual}) and hydraulics modeling of WDNs (Section~\ref{sec:foundations:hydraulics}), including the types of events in WDNs that we consider in this work. Next (Section~\ref{sec:problemformulation}), we introduce the event diagnosis interpretability problem for which we propose a solution in this work -- the counterfactual event fingerprints, as our key contributions, are proposed in Sections~\ref{sec:problemformulation} and~\ref{sec:fault_isolation}. After these conceptional contributions, we evaluate the performance of our proposed methodology in two case studies (Sections~\ref{sec:experiments}-\ref{sec:experiments2}).
Finally, this work closes with a summary and conclusion in Section~\ref{sec:conclusion}.

\section{Foundations of Counterfactual Explanations}\label{sec:foundations:counterfactual}
Contrasting explanations are a popular explanation method in XAI~\cite{molnar2019} because there exists strong evidence that explanations by humans (which these methods try to mimic) are often contrasting in nature~\cite{CounterfactualsHumanReasoning}; i.e., people  ask questions like \textit{``What needs to be changed in order to observe a different outcome?''}.
A prominent instance of contrasting explanations are counterfactual explanations~\cite{CounterfactualWachter,CounterfactualReviewChallenges,guidotti2022counterfactual} (often just called \textit{counterfactuals}), which aim to derive a change to some features of a given input such that the resulting data point, called the counterfactual, causes a different behavior of the system compared to the original input. Thus, one can think of a counterfactual explanation as a suggestion of actions (changes to the input) that change the model's behavior (output).
For illustration purposes, consider the example of loan application: \textit{Imagine an individual who applied for a loan at a bank. Unfortunately, the bank rejected the application. Now, the individual would like to know why. In particular, the individual would like to know what would have to be different so that their application would have been accepted. A possible explanation might be that the application would have been accepted if they had earned 500\$ more per month and if they had not had a second credit card.}

In order to keep the explanation simple and easy to understand (low-complexity explanation), an obvious strategy is to aim for a ``minimum amount of change'' in the input features to achieve the desired output. This is formalized in the following definition.
\begin{definition}[(Closest) Counterfactual Explanation~\cite{CounterfactualWachter}]\label{def:counterfactual}
	Assume a prediction function $\classifier:
	\setX\to \setY$ is given, where $\setX\subseteq\RN^\dimsym$ and $\setY\subseteq\RN^q$. A counterfactual $\xcf \in \setX$ for a given input $\xorig \in \setX$ is given as a solution to the following optimization problem:
	\begin{equation}\label{eq:counterfactualoptproblem}
		\underset{\xcf \,\in\, \setX}{\arg\min}\;\left\{ \loss\big(\classifier(\xcf), \ycf\big) + C \cdot \regularization(\xcf, \xorig)\right\}
	\end{equation}
	where $\loss(\cdot)$ denotes a pre-selected loss function, $\ycf\in\setY$ is the desired target prediction, $\regularization(\cdot)$ denotes a penalty for dissimilarity of $\xcf$ and $\xorig$, and $C>0$ is a constant that characterizes the regularization strength.
\end{definition}
Such counterfactuals are also called \textit{closest counterfactuals} because they aim for the closest (i.e.\ most similar) explanation $\xcf$ to the input $\xorig$. However, other desirable aspects like plausibility and actionability are not included in~\refdef{def:counterfactual}. For example, it was observed~\cite{poyiadzi2020face} that in some applications such counterfactuals can be highly implausible (e.g.\ looking like adversarials~\cite{chakraborty2021survey}) or the suggested actions may not be possible to execute in real life (e.g.\ changing the age in a loan application). Such additional aspects and constraints are investigated in works such as~\cite{ActionableCounterfactuals,CounterfactualGuidedByPrototypes,PlausibleCounterfactuals}.

Many explanation methodologies, including contrasting explanations like counterfactual explanations, suffer from non-uniqueness, in the sense that there may exist more than one valid explanation -- this is called the ``Rashomon effect''~\cite{molnar2019}. In such cases, it is not clear which or how many counterfactual explanations should be presented to the user. One common modeling approach is to compute a set of maximally diverse explanations instead of a single explanation~\cite{mothilal2020explaining,russell2019efficient,rodriguez2021beyond,artelt2022explaindimred}, or to add a suitable regularization to enforce uniqueness.

Another limitation of existing explanation methodologies (including counterfactual explanations) is that they cannot be applied to an ensemble of decisions~\cite{CounterfactualReviewChallenges} -- in other words, existing methods can only explain a single outcome but not multiple ones like it is the case in an ensemble of models that have a large overlap of the inputs they process.
In this context, a key contribution of the present work is the proposal of a novel explanation concept called \textit{counterfactual fingerprint} (see Section~\ref{sec:counterfactual_event_detection_fingerprint}) for explaining the outputs of a set of virtual sensors deployed on the same WDN. These \textit{counterfactual fingerprints} are a crucial building block of our proposed interpretable event diagnosis methodology.

\section{Hydraulics and Fault Modeling of WDN}\label{sec:foundations:hydraulics}
The topology of a WDN is modeled by a directed graph denoted as $\mathcal{G}=(\mathcal{N},\mathcal{L})$. 
Let $\mathcal{N} = \left\{ {1, \cdots ,{n_n}} \right\}$ be the set of all nodes, where $\left| \mathcal{N} \right| = n_n$ is the total number of nodes.
These represent junctions of pipes, consumer water demand locations, reservoirs, and tanks.
Moreover, let $\mathcal{N}_r \subset \mathcal{N}$ represent the subset of reservoir and tank nodes.
The hydraulic state associated with each node $j\in \mathcal{N}$ is the \textit{hydraulic head} $h_j(t)$, where $t$ is the (continuous) time notation for variables.
The hydraulic head consists of a component analogous to the pressure $p_j$ at node $j$, and of the node elevation $ z_j $ with respect to a geodesic reference.
Each node $j$ is also associated with a water consumer demand at the node location, denoted by $d_j(t)$.
Water demands at each node drive the dynamics of a WDN and are typically unknown, even though the aggregate demands are partially known.
Let $\mathcal{L} = \left\{ {1, \cdots ,{n_l}} \right\}$ be the set of links, where $\left| \mathcal{L} \right| = n_l$ is the total number of links.
These represent network pipes, water pumps, and pipe valves.
The hydraulic quantity associated with a link $i \in \mathcal{L}$ is the \textit{water flow}, indicated by $q_i(t)$. 

The hydraulic equations that describe the behavior of the water distribution system are the conservation of energy equations and the conservation of mass equations.
Energy in WDN is associated with the head at nodes.
The head difference between two network nodes which are connected by a link $i$ through which there is water flow $q_i$, is given by a nonlinear function $f_i(u_i(t),q_i(t))$. 
The control input $u_i(t)$ concerns links $i$ that correspond to actuators of a WDN, namely pumps and valves.
We assume that $u_i=0$ for all other links, thus the control input vector is given by $u(t) \in \mathbb{R}^{n_c}$, with $n_c$ the number of actuators. 
The \emph{energy equations} for network link $i$, can be written as follows \cite{Vrachimis2018a}:
\begin{equation}\label{eq:energy_balance}
		f_i\left(u_i(t), q_i(t)\right) + \sum\limits_{j \in \mathcal{N}} {\left(B_{ij}~{h_j(t)}\right)} =  0 ,~~~ i \in \mathcal{L},
\end{equation}
where 
${B} \in {\mathbb{R}^{{n_l} \times {n_n}}}$ is the incidence flow matrix, indicating the connectivity of nodes with links such that
element $B_{ij}=+1$ if the conventional direction of link $i$ enters node $j$;
element $B_{ij}=-1$ if the conventional direction of link $i$ leaves from node $j$;
otherwise  $B_{ij}=0$. 
Each function $ f_i(u_i(t),q_i(t)) $ represents the head-loss at link $i\in \mathcal{L}_p$, which is a measure of the energy dissipated due to friction of water flowing through the link.
The head-loss function $ f_i(u_i(t),q_i(t)) $ can be selected to represent system components modeled at links, such as pipes, pumps, and valves \cite{Vrachimis2018a}.
In the case of pipes, head-loss depends on the water flow through the pipe but also on pipe parameters, such as pipe length, diameter, and pipe roughness coefficient.

The conservation of mass law for each node $j \in \mathcal{N}$ dictates that the sum of branch water flows from pipes incident to a node $j$ must be equal to the node's external outflows.
The \textit{conservation of mass equations} for a node $j$, can be written as follows \cite{Vrachimis2018a}:
\begin{equation}\label{eq:node_balance}
\sum\limits_{i \in\mathcal{L}} {\left(B^\top_{ij}\ {q_i(t)}\right)}  = d_j(t)  + {\xi_j^l(t,h_j(t))},	 j \in \mathcal{N} 
\end{equation}
where $\xi^l_j$ represents a pressure-dependent leakage occurring at node $j$.
The underlying modeling assumptions in \eqref{eq:node_balance} are: 
i) The hydraulic heads at reservoir and tank nodes are measured and known;
ii) The network operates in pressure-sufficient conditions, such that the demand at each node is independent of the node pressure.

The complete hydraulic state of a WDN, comprised of the water flows in pipes and hydraulic heads at nodes, is denoted by 	
\begin{equation}
x(t) = \begin{bmatrix}
q(t) \\
h(t)
\end{bmatrix} \in \mathbb{R}^{n_l+n_n}.
\end{equation}

Next, we consider abnormal events that may occur in WDNs.
In this work, we focus on two types of events that may occur in a WDN: Leakages and faults in hydraulic sensors.
These are indicated by the function $\xi^F_j(t)$, where $F\in \{l,s\}$ denotes water leakages and sensor faults, respectively.
The leaks in this work are modeled on network nodes, with $\xi^l_j(t)$, $j\in \mathcal{N}$ corresponding to the index of the node on which the leak has occurred.
For sensor faults, $\xi^s_j(t)$, $j\in \{1,..,n_s\}$ corresponds to the index of the sensor that is experiencing a fault.
Other hydraulic events that may also occur in the network, such as actuator faults (valves, pumps) and abnormal water consumption by consumers, will be considered in future work.

A leak affects the dynamics of the network as described by~\eqref{eq:node_balance} with the leakage magnitude assumed to be pressure dependent. 
The pressure-dependent leakage function \cite{Kabaasha2020} is given by:
\begin{equation}\label{eq:leakage_function}
    \xi^l_j(t,h_j(t)) = c_j(t) (h_j(t)-z_j)^\alpha ,
\end{equation}
where $c_j(t)\geq0$ is the time-dependent leak emitter coefficient and $\alpha>0$ is the leak emitter exponent.

In the case of a typical District Metered Area (DMA) of a WDN, measurements are available from sensors at discrete time steps which correspond to a time interval $\Delta t$.  We refer to \emph{observed time points} as $k$ to emphasize the difference from underlying continuous time $t$. 
Typically flow is measured at the inlets; flow measurements are denoted by $y^q(k) \in \mathbb{R}^{n_i}$, where $n_i$ is the number of inlets.
Additionally, some pressure sensors may be installed inside the DMA, giving pressure measurements indicated by $y^p(k) \in \mathbb{R}^{n_s}$, where $n_s$ is the number of pressure sensors. 
Sensor measurements are typically noisy, whereby the noise is random and varies depending on the type of device.
Moreover, there may be unexpected sensor faults, which may occur at an unknown time $t$, causing errors in the measurements. 
The true value of the measured state relates to the measurements as follows:
\begin{align}\label{eq:measurements}
y(k) = \begin{bmatrix}
{y}^q(k)\\
{y}^p(k)
\end{bmatrix}
 = C_y 
\begin{bmatrix}
{q}(k) \\
{h}(k) - {z}
\end{bmatrix}
 + \begin{bmatrix}
 {{e}}^q(k)\\
{{e}}^p(k)
\end{bmatrix}
+ {\xi}^s(k,x)
\end{align}
where  
$C_y \in \mathbb{R}^{(n_i+n_s)\times(n_l+n_n)}$ is a diagonal matrix used to identify measured flow and head states, ${z}\in \mathbb{R}^{n_n}$ is the vector of node elevations, and $e^q$, $ e^p $ is the measurement error for flow and pressure devices respectively.
Moreover, ${\xi}^s \in \mathbb{R}^{n_i+n_s}$ represents a vector of various types of sensor faults which may cause sensor measurements to be inaccurate.

Individual sensor faults are represented by $\xi^s_i(k,x)$ where the subscript $i$ corresponds to the sensor associated with the fault (e.g., the $i$-th sensor), while $k$ and $x$ indicate that the fault may be time and state-dependent respectively \cite{Reppa2016}.

\section{Interpretable Event Diagnosis in WDNs}\label{sec:problemformulation} 

Generally, \emph{Fault Detection} refers to methodologies that aim to give an alert $\mathcal{A}(k) \in \{0,1\}$ when, provided with sensor measurements, a 
fault is detected
\cite{Isermann2006}.
The fault detection methodology can be purely data-driven or rely on a model of the examined system.

\emph{Fault Isolation} refers to methodologies that, given the detection of a fault, are able to determine the fault type $F$ and/or its location $j$.

The objective of this work is to develop a purely data-driven fault detection and isolation methodology, through event classification, that is \textit{interpretable}, i.e., the results are accompanied by an explanation that the system operators can associate with their experience.

In Section~\ref{sec:faultdetection}, we will present specific data-driven instantiations of fault detection and fault isolation in WDN, which are based on sensor forecasting and residual methods. Before, we present a general scheme of how such detection and isolation methods can be designed.
We propose that insight can be gained into the data-driven methodology by providing contrasting explanations of the outputs -- i.e., alternative results when the inputs are varied.
For this purpose, the corresponding explanation components can be associated with any functions which are used for fault detection and/or isolation.

\begin{figure}
	\centering
	\includegraphics[scale=1.25]{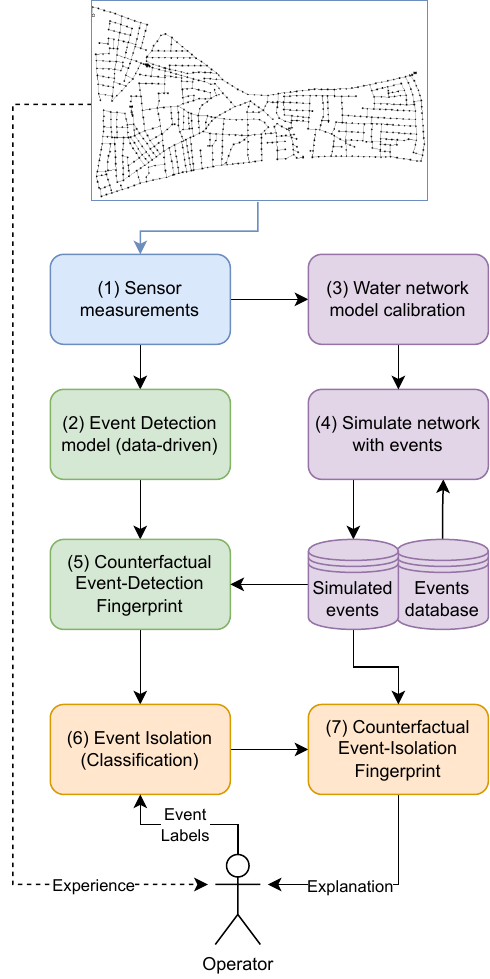}
	\caption{Illustration of the proposed Interpretable Event Diagnosis methodology: The blue block indicates the sensor component for collecting data, the purple blocks indicate the modeling component for the WDN, the green blocks indicate the event detection components, and the orange blocks indicate the event isolation components.}
 	\label{fig:methodology}
\end{figure}

Our proposed framework for tackling the event diagnosis interpretability problem in WDN is illustrated in Fig.~\ref{fig:methodology}; it comprises of the following steps:
\begin{enumerate}
	
	\item Collect historical measurement data from the real network for a time period that captures the periodic behavior of the system under normal conditions.
	
	\item Data-driven event detection: Using the real sensor data, build a data-driven event detection system by training a machine learning model to create virtual sensors that are able to estimate the pressure at a sensor node, using measurements from other sensor nodes.

	\item Calibrate a digital model of the real network (using the real data) in which the normal operation corresponds to the real data gathered. 
	
	\item Create a database of a simulated set of events on the digital model, such as leakages of different magnitudes/locations and sensor faults of different types. 
	
	\item Use this event detection system for producing \textit{counterfactual event detection fingerprints} (CDF) of the detected events -- i.e. explaining the event detection: What are feasible changes to the inputs (i.e. sensor measurements) such that the event would not have been detected?
	
	\item Event isolation: Train a classification algorithm to distinguish the type of event from their CDFs -- the classifier is trained based on the simulated events from the digital model. 
	
	\item Produce \textit{counterfactual event isolation fingerprint} (CIF) -- i.e. explaining the event isolation: What are feasible changes to the counterfactual event detection fingerprint of the event that would result in a different classification of this event?
	
	\item Monitor the real system until an event is detected, classify the event, produce explanations, and receive feedback from the operator, e.g., a label indicating the type of the observed event based on their experience.
\end{enumerate}

We will now introduce specific methods for steps (2) and (6), and introduce the formal definition of CDF and CIF as counterfactual explanations associated with these two mappings.
To demonstrate this methodology, we provide a case study based on the Hanoi network \cite{Vrachimis2021} in Section~\ref{sec:experiments}, and another one based on the realistic L-Town benchmark~\cite{Vrachimis2022} in Section~\ref{sec:experiments2}.

\FloatBarrier

\section{Data Driven Event Detection in WDNs}\label{sec:faultdetection}
In this section, we describe a machine-learning-based data-driven event detection methodology which we evaluate in our case study in Section~\ref{sec:experiments}.
We assume that we are operating a WDN with $\numSensors$ pressure (and optional additional flow) sensors. Furthermore, we assume that we are given a data stream of sensor values.
Formally, a fault detection function  can be represented by a generic function that acts on a time window of $T$ sensor signals (we ignore possible additional control signals, for simplicity):
$f_D\left(\cdot\right) : \mathbb{R}^{n_s\times T}\to \{0,1\}$ 
giving output $f_D(\cdot)=1$ iff a fault is present at the current point in time.

\paragraph*{Implementation of event detection:} We realize $f_D(\cdot)$ via a classic two-step residual-based event detection system as follows:
In the first step, for each sensor, a virtual copy is instantiated at the same location; the value of sensor $i$ is predicted based on the matrix of the last $T$ measurements of all other sensors (pressure and flow sensors) referred to as $\mat{Y}$, using a trainable function $f_i:\RN^{\numSensors\times T} \to \RN$. 
The sensor forecasts $f_i(\cdot)$ (virtual sensors for event detection) are realized in this work using linear regression, as follows:
\begin{equation}
	f_i(\mat{Y}_{[\timeVar-T-1,\ldots,\timeVar-1]}) = \vec{w}_i^\top\frac{1}{T}\sum_{j=1}^T \vec{y}({\timeVar-i)} + b_i
\end{equation}
where 
$\mat{Y}_{[\timeVar-T-1,\ldots,\timeVar-1]}$
refers to the time window of the last $T$ sensor values. 
The vector notation $\vec y(k)$ refers to the sensor values at one time step $k$.
Here we take the average of the past sensor measurements over a time window of size $T$ as a feature vector for a linear regression.
The parameters $\vec{w}_i$ and $b_i$ are estimated on a part of the data stream for every $i$, where we know that no fault is present, using a (regularized) least-squares estimation. The entry $i$ of vector $\vec w_i$ is set to $0$ such that sensor values are inferred based on all other measurements ignoring the sensor $i$ itself.

Note that other approaches can be used to generate sensor forecasts such as nonlinear graph models \cite{DBLP:conf/aaai/AshrafSHH24}; however, we choose linear regression because it is a relatively simple approach that works well on this application.
A combination of these individual functions per sensor yields the forecast $f:\RN^{\numSensors\times T} \to \RN^{\numSensors}$ with vector of sensor values
$f(\mat{Y}_{[\timeVar-T-1,\ldots,\timeVar-1]})$.

The event detection itself is then implemented by applying a threshold to the residuals -- i.e. considering the difference between the forecasts and observed pressure values at all available virtual sensors. An alarm is raised if at least one of the differences exceeds a given threshold:
\begin{equation}\label{eq:residual_event_detection}
	\pnorm{f(\mat{Y}_{[\timeVar-T-1,\ldots,\timeVar-1]}) - \vec{y}({\timeVar})}_{\infty} > \theta
\end{equation}
where $\vec{y}(\timeVar)$ denotes the sensor measurements observed at time $\timeVar$.
The threshold $\theta > 0$ is also estimated on a period without any faults -- it might be chosen to be a bit larger than the maximum observed residual in this training period. Instead of a global threshold, sensor-specific thresholds $\theta_i$ for $f_i(\cdot)$ can be used.
This event detection methodology can be rephrased as event detection function  function $f_{D}(\cdot),$ via the indicator function
$f_D(\mat{Y}_{[\timeVar-T-1,\ldots,\timeVar-1]})=
\mathbf{1}\left(\pnorm{f(\mat{Y}_{[\timeVar-T-1,\ldots,\timeVar-1]}) - \vec{y}({\timeVar})}_{\infty} > \theta\right)
$

\begin{figure}[ht!]
	\centering
	\includegraphics[width=\textwidth]{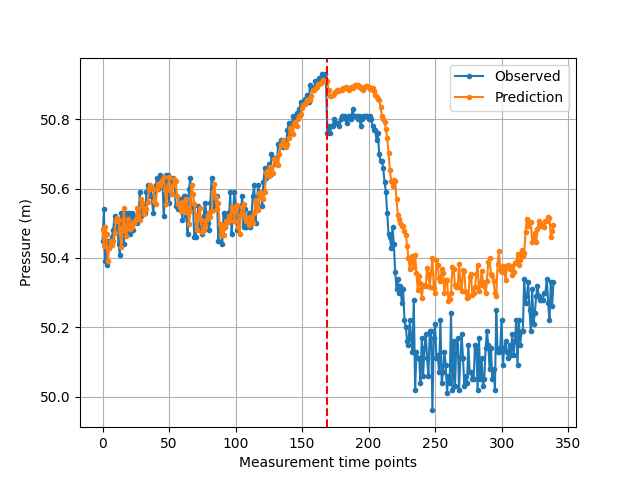}
	\caption{L-Town-Network ($\Delta t = 5min$): Pressure forecasts in normal vs. leaky times -- a leakage is present from time $170$ onwards (starting point indicated by the vertical dashed red line).}
\label{fig:pressure_forecast}
\end{figure}

For illustrative purposes, we show a short window of a non-faulty time period followed by a faulty (a leakage is present) time period in Fig.~\ref{fig:pressure_forecast}.

\section{Counterfactual Event Detection Fingerprints (CDF)}\label{sec:counterfactual_event_detection_fingerprint}
In the following, we formalize the concept of \textit{counterfactual event detection fingerprints} which are used to explain the event detection and, as a downstream task, to classify (i.e.\ determine the type of) the observed event itself.
\begin{definition}
Given an event detection method characterized by the function $f_D:\RN^{n_s\times T}\to\{0,1\}$, and a specific time step $k$ where an alert is raised, i.e., 
$f_D(\mat{Y}_{[\timeVar-T-1,\ldots,\timeVar-1]})=1$; a \textit{Counterfactual event Detection Fingerprint} (CDF) $\delta_{CDF}(k)\in\RN^{n_s\times T}$ consists of vectors of sensor measurements that, when added to the actual sensor measurements in the observed time window,
 would cause a change of the alert signal from $1$ to $0$, i.e., no event would be detected.
 Thus it is characterized by the property
\begin{equation}\label{eq:counterfactual_event_detection_fingerprint:pred}
f_D(\mat{Y}_{[\timeVar-T-1,\ldots,\timeVar-1]}+\delta_{CDF}(k))=0
\end{equation}
The set of counterfactual fingerprints fulfilling this property is referred to as $\Delta_{CDF}(k)$
\end{definition}
Since our specific implementation of $f_D(\cdot)$ depends on 
an ensemble of predictions for each individual sensor,
 we have to make sure that a CDF fits all functions $f_i(\cdot)$ in the ensemble of virtual sensors.
 
CDFs $\Delta_{CDF}(k)$ in \refeq{eq:counterfactual_event_detection_fingerprint:pred} are not unique. 
As it is common practice in XAI~\cite{molnar2019}, we add a regularization by aiming for the ``simplest'' possible CDF -- i.e., a fingerprint $\delta_{CDF}(k)\in\Delta_{CDF}(k)$  where as few sensor values as possible are changed (i.e., regularization w.r.t.\ zero-norm $\pnorm{\cdot}_0$).
With this in mind, we can phrase the computation of a counterfactual event detection fingerprint as ensemble consistent explanation~\cite{artelt2022ijcai} -- i.e., a low-complexity (simple) explanation that explains all functions $f_i(\cdot)$ in the ensemble: 
\begin{definition}[Closest Counterfactual Event Detection Fingerprint]\label{def:closest_counterfactual_event_detection_fingerprint}
The \textit{closest counterfactual event detection fingerprint} (closest CDF) $\delta^*_{CDF}(k)\in\Delta_{CDF}(k)$ of an ensemble of event detection functions $f_i(\cdot)$ is given as the solution to the following optimization problem:
\begin{equation}\label{eq:cffingerprint:opt}
	\begin{split}
		&\underset{\delta^*_{CDF}(k)\,\in\,\Delta_{CDF}(k)}{\min}\,\pnorm{\delta^*_{CDF}(k)}_0 \quad
		\text{s.t. }  \Big|\regression_i\left(\mat{Y}_{[\timeVar-T-1,\ldots,\timeVar-1]} + {\delta^*_{CDF}}(k)\right) - {\y}(k)_i\Big| \leq \theta \quad \forall\,i\\
	\end{split}
\end{equation}
\end{definition}
While~\refeq{eq:cffingerprint:opt} from~\refdef{def:closest_counterfactual_event_detection_fingerprint} can be solved to compute a CDF, it can not be guaranteed that this resulting CDF is reasonable or plausible in the domain -- e.g., negative pressure values might result. Therefore, additional constraints might be incorporated to get a reasonable CDF of the event. A straightforward approach is to limit the set of possible solutions (i.e., CDFs $\Delta_{CDF}(k)$) to a previously recorded set of measurements that do not lead to alarms.

\paragraph*{Implementation}
As this will be sufficient in practice,  we set the length of the time interval $T$ to $1$ (i.e., only a single point in time is considered). Extensions of the computational approach to longer time intervals are immediate.
In order to make the computation of a counterfactual event detection fingerprint (\refdef{def:closest_counterfactual_event_detection_fingerprint}) tractable, we relax the objective of minimizing the number of changes~\refeq{eq:cffingerprint:opt} and use the $1$-norm ($\pnorm{\cdot}_{1}$) as an approximation. Furthermore, in order to guarantee plausibility of the final closest CDFs $\delta^*_{CDF}(k)$ (\refdef{def:closest_counterfactual_event_detection_fingerprint}), we limit the set of data which are obtained by feasible (i.e., possible) CDFs $\Delta_{CDF}(k)$ to the set of training samples $\set{D}$. This is reasonable as we fit the event detection method to the set of measurements $\set{D}$ from a time period without any events (e.g., faults).
Hence, denoting 
$\vec x_{cf}=\vec{y}(k-1)+\delta^*_{CDF}(k)$ 
the problem reduces to the challenge of selecting the most similar sensor measurements from the training set $\xcf\in\set{D}$ which fulfills the following property:
\begin{equation}\label{eq:explanation:fault_detection}
	\underset{\xcf\,\in\,\set{D}}{\min}\,\pnorm{\xcf - \vec{y}(\timeVar-1)}_1 \quad
		\text{s.t. } \Big|\regression_i\left(\xcf\right) - {\y}({\timeVar})_i\Big| \leq \theta_i \quad \forall\,i
\end{equation}
where ${y}({\timeVar})_i$ denotes the observed pressure at the $i$-th sensor at time $\timeVar$. Note that, as before, $f_i$ does not use 
the $i$-th sensor for prediction.
The closest CDF itself $\delta^*_{CDF}(k)$ is given as the difference 
\begin{equation}
    \delta^*_{CDF}(k) = \xcf - \vec{y}({\timeVar-1})
\end{equation}

While a sensor fault affects a single sensor only, a leakage might affect multiple sensors.
Therefore it is not surprising that we consistently observe significant differences in the obtained CDFs for leakages and sensor faults -- the existence of these differences will allow 
us to build a reliable and interpretable classifier for event isolation (see Section~\ref{sec:fault_isolation}), which is based on CDFs. This will be based on the following observation: While in case of a leakage, a lot of affected nodes are highlighted in the CDF, few, in the limit only one (the faulty) sensor are highlighted in the CDF in case of a sensor fault -- see Fig.~\ref{fig:hanoi:fingerprint:examples} from the first case study (Section~\ref{sec:experiments}) for an example illustration.

\section{Event Isolation (Classification)}\label{sec:fault_isolation}
Fault classification refers to the challenge, given that an event has been detected, to identify the type of the fault.
For simplicity, we restrict to the setting of one fault, which corresponds to either leakage (class $0$) or sensor fault (class $1$); computational extensions to more types would be straightforward.
The fault identification corresponds to a generic function 
 $f_{I}\left(\cdot\right) : \mathbb{R}^{n_s\times T}\to \setY=\{0,1\}$ that returns the fault type within the set of possible faults $\setY$, given that a detection occurred. 

\paragraph*{Implementation}
We classify the observed event based on their CDF using a standard classifier such as a k-nearest neighbor classifier or a decision tree classifier. For this purpose, we build and simulate a set of scenarios (see the two case studies in Sections~\ref{sec:experiments}-\ref{sec:experiments2}) where we either have a leakage or a sensor fault.
We apply the fault detection method and compute the closest CDFs $\delta^*_{CDF}(k)$ of the raised alarms as introduced in \refdef{def:closest_counterfactual_event_detection_fingerprint}. This way we obtain a labeled set of CDFs of $L$ representative leakages ($y=0$) and sensor faults ($y=1$) of the observed system at hand, purely from simulations:
\begin{equation}\label{eq:event_isolation:trainingset}
    \set{D}_{train} = \{{(\delta^*_{CDF}(k)}_l, y_l)\:|\:l=1,\ldots,L\} \quad\text{with } y_l \in \{0, 1\}
\end{equation}
We use this training set~\refeq{eq:event_isolation:trainingset} for training a classifier (e.g.\ k-nearest neighbor or decision tree classifier) $\classifier: \RN^{n_s} \to \{0, 1\}$ that can distinguish between sensor faults and leakages by looking at the given CDF. Since CDFs are interpretable and methods such as $k$-NN and decision trees are interpretably by design, the entire classification becomes interpretable.

\paragraph*{Non-interpretable baseline}\label{sec:event_isolation:baseline}
In order to compare our proposed interpretable event isolation methodology (see previous section), we compare the result to an alternative, which directly classifies the sensor residuals $\Delta\y=\mathrm{abs}(f(\vec y(k-1))-\vec y(k))\in\RN^{n_s}$ as a baseline where $\mathrm{abs}$ refers to the component-wise absolute value -- i.e.\ we train a classifier to classify the event based on the
values $\Delta\y$ and associated label information.
Similar to our proposed interpretable event isolation method, we implement the classifier as a decision tree classifier and as a k-nearest neighbor classifier.
Besides possible (minor) differences in the performance of the baseline and the interpretable event classifier, we expect significant differences in their interpretability.

\section{Counterfactual Event Isolation Fingerprints}\label{sec:counterfactual_isolation_fingerprint}
Similar to the problem of fault detection (see Section~\ref{sec:counterfactual_event_detection_fingerprint}), we do not have access to the true fault isolation function but only to an estimate $\classifier(\cdot)$ -- see Section~\ref{sec:fault_isolation}. 
Therefore, we propose an auxiliary inspection technology that, by means of counterfactual explanations, enables practitioners to 
inspect alternative sources of the observed fault.
For this purpose, we introduce the general notion of Counterfactual Event Isolation Fingerprints, which indicate the
characteristics responsible for the observed fault as opposed to alternatives:

\begin{definition}[Counterfactual Event Isolation Fingerprint]\label{def:cif}
The \textit{Counterfactual event Isolation Fingerprint} (CIF) $\delta_{CIF}(k) \in\RN^{n_s\times T}$ is any vector that, given a fault has been detected, when added to 
a pre-calculated CDF $\delta^*_{CDF}(k)$  would cause the fault type to change, e.g., the classification changes from a sensor fault to leakage.
This implies the inequality 
\begin{equation}\label{eq:counterfactual_event_isolation_fingerprint:pred} 
 o(\delta^*_{CDF}(k)) \quad\neq\quad  o(\delta^*_{CDF}(k)+\delta_{CIF}(k))
 \end{equation}
\end{definition}
The set of all possible CIFs for which \refeq{eq:counterfactual_event_isolation_fingerprint:pred} holds at time $k$ is denoted by $\Delta_{CIF}(k)$.

There exists multiple possible CIFs $\Delta_{CIF}(k)$ (\refdef{def:cif}) -- i.e. multiple explanations of the event isolation. We propose to select the ``simplest'' CIF $\delta^*_{CIF}(k)$ -- i.e., select the CIF $\delta^*_{CIF}(k)\in\Delta_{CIF}(k)$ with the smallest number of changes:
\begin{definition}[Closest Counterfactual Event Isolation Fingerprint]\label{def:closest_cif}
The closest counterfactual event isolation fingerprint (closest CIF) $\delta^*_{CIF}(k)$ of a given closest CDF $\delta^*_{CDF}(k)$ is given by the solution to the following optimization problem: 
\begin{equation}\label{eq:closest_counterfactual_event_isolation_fingerprint:opt}
	\begin{split}
		&\underset{\delta^*_{CIF}(k)\,\in\,\Delta_{CIF}(k)}{\min}\,\pnorm{\delta^*_{CIF}(k)}_0 \quad
		\text{s.t. } \classifier(\delta^*_{CDF}(k) + \delta^*_{CIF}(k)) \neq \classifier(\delta^*_{CDF}(k))
	\end{split}
\end{equation}
\end{definition}
Similar to the counterfactual event detection fingerprints (Section~\ref{sec:counterfactual_event_detection_fingerprint}), we can not guarantee the plausibility of the solutions to~\refeq{eq:closest_counterfactual_event_isolation_fingerprint:opt}. As before, a straightforward approach is to limit the feasible set $\Delta_{CIF}(k)$ to previously recorded CDFs.

\paragraph*{Implementation}
We implement the computation of a closest counterfactual event isolation fingerprint $\delta^*_{CIF}(k)$ (see~\refdef{def:closest_cif}) in a manner that closely parallels the computation of a closet CDF (\refdef{def:closest_counterfactual_event_detection_fingerprint}): We use the $1$-norm as the objective and limit the set of feasible solutions $\Delta_{CIF}(k)$ to a set of previously recorded CDFs $\set{D}_{CDF}$~\refeq{eq:event_isolation:trainingset} -- as an example, this could be chosen as the training set $\set{D}_{CDF}=\set{D}_{train}$ used to train the event classifier $\classifier(\cdot)$. Then the optimization problem becomes as follows:
\begin{equation}
	\begin{split}
		&\underset{\xcf\,\in\,\set{D}_{CDF}}{\min}\,\pnorm{\xcf - \delta^*_{CDF}(k)}_1 \quad
		\text{s.t. } \classifier(\xcf) \neq \classifier(\delta^*_{CDF}(k))
	\end{split}
\end{equation}
The final CIF $\delta^*_{CIF}(k)$ is then given as follows:
\begin{equation}
    \delta^*_{CIF}(k) = \xcf - \delta^*_{CDF}(k)
\end{equation}

\section{Case Study I}\label{sec:experiments}
\begin{figure}[h!]
	\centering
    \includegraphics[width=0.8\linewidth]{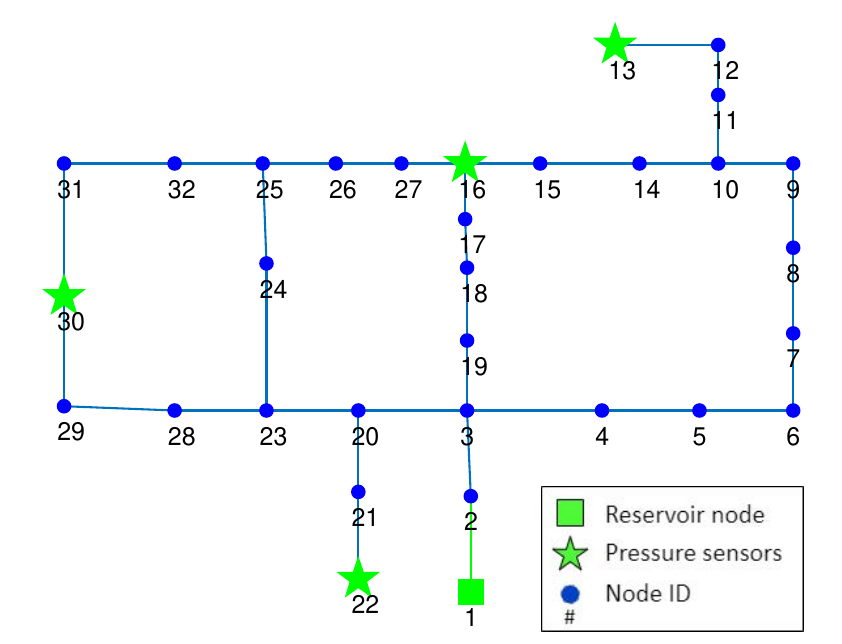}  
	\caption{The Hanoi network including four ($4$) pressure sensors and one ($1$) flow sensor at the inlet.}
 	\label{fig:Hanoi_sensors}
\end{figure}

First, we implement and apply our proposed methodology on a simple benchmark water transport network where we encounter two types of events: leakages and sensor faults.
In particular, we consider the following four types of sensor faults~\cite{Reppa2016}:
\begin{enumerate}
	\item Gaussian noise added to the sensor measurements, with a significantly larger standard deviation than sensor specifications.
	\item Sensor power failures, which result in the measurement being equal to zero.
	\item An offset of the sensor measurement, linearly proportional to the true measured quantity.
	\item Sensor drift, resulting in an increasing measurement value until a maximum limit is reached.
\end{enumerate}
The ``Hanoi'' benchmark network, shown in Fig.~\ref{fig:Hanoi_sensors}, is a transport network where each node represents a District Metered Area with unique realistic demand patterns exhibiting a $24$ hour periodicity.

A version of the network, which serves as the ``real'' system in our experiments, is simulated for $48$ hours, with a hydraulic time-step of $\Delta t = 30$ minutes.
The ``real'' system's inlet is monitored, i.e., the head at the reservoir node $1$ and the flow at link $1$ are measured.
Additionally, $4$ pressure sensors are installed at nodes $ \{13,16,22,30\} $ using a sensor placement procedure that maximizes the minimum sensitivity of all sensors to all possible leakages \cite{Casillas2013}.

The Event Detection module (see Section \ref{sec:faultdetection}) is trained using data from these sensors during the network's normal operation.

Next, a nominal hydraulic model of the network is given which is assumed topologically identical to the real network.
It is assumed that this nominal model has been calibrated using sensor data from the ``real'' network.
Since the calibration procedure does not yield accurate results, the nominal pipe parameters are within $\pm 5\%$ of the parameter values of the ``real'' network. 
Moreover, demands are within $\pm 10\%$ of the ``real'' demand values.
The nominal model is used to train the event classifier, by simulating $204$ scenarios with a single leak (we vary time, size, and location) and $52$ scenarios with a single sensor fault (we vary time, magnitude, type, and location) -- all scenarios are three months long and they have the same number of scenarios with a small, medium or large leakage.

\subsection{Illustrative Examples}
\begin{figure}
	\centering
	\begin{minipage}[b]{0.49\textwidth}
		\includegraphics[width=\textwidth]{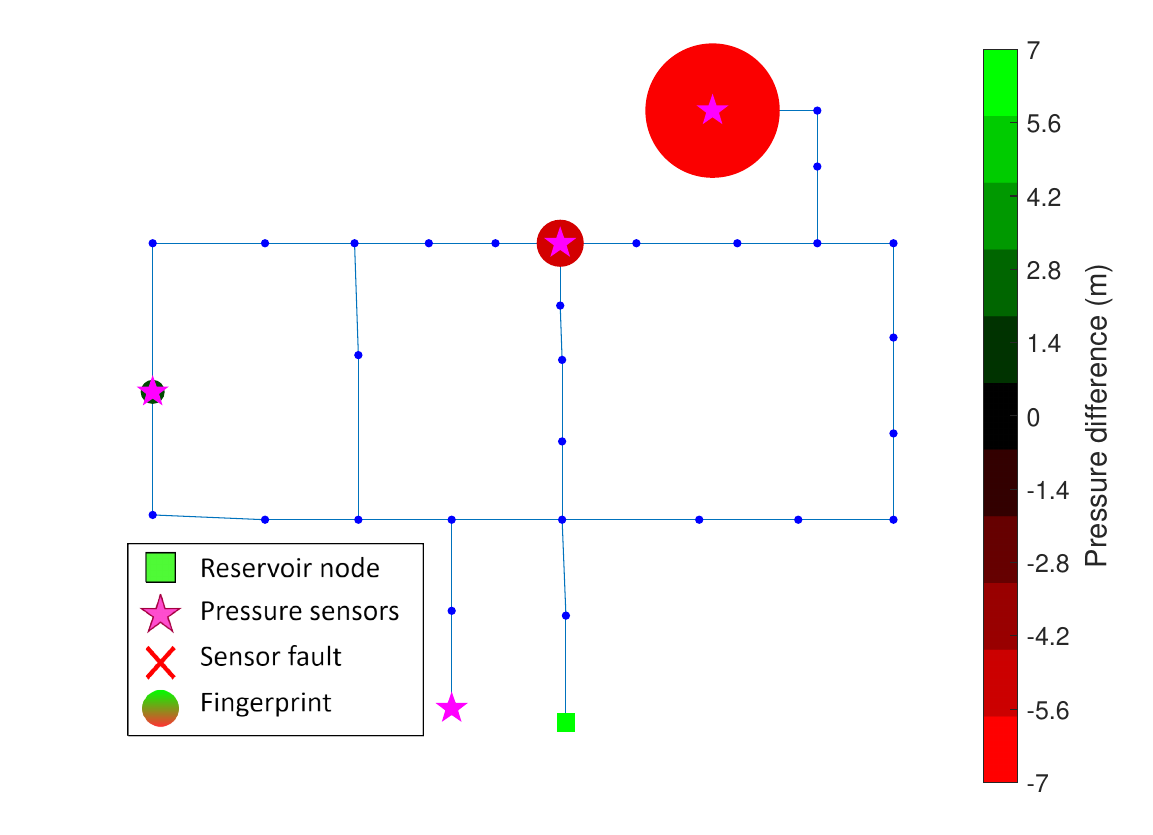}  
		\caption*{(a) Sensor fault at sensor 13.} 
	\end{minipage}
	\hfill
	\begin{minipage}[b]{0.49\textwidth}
		\includegraphics[width=\textwidth]{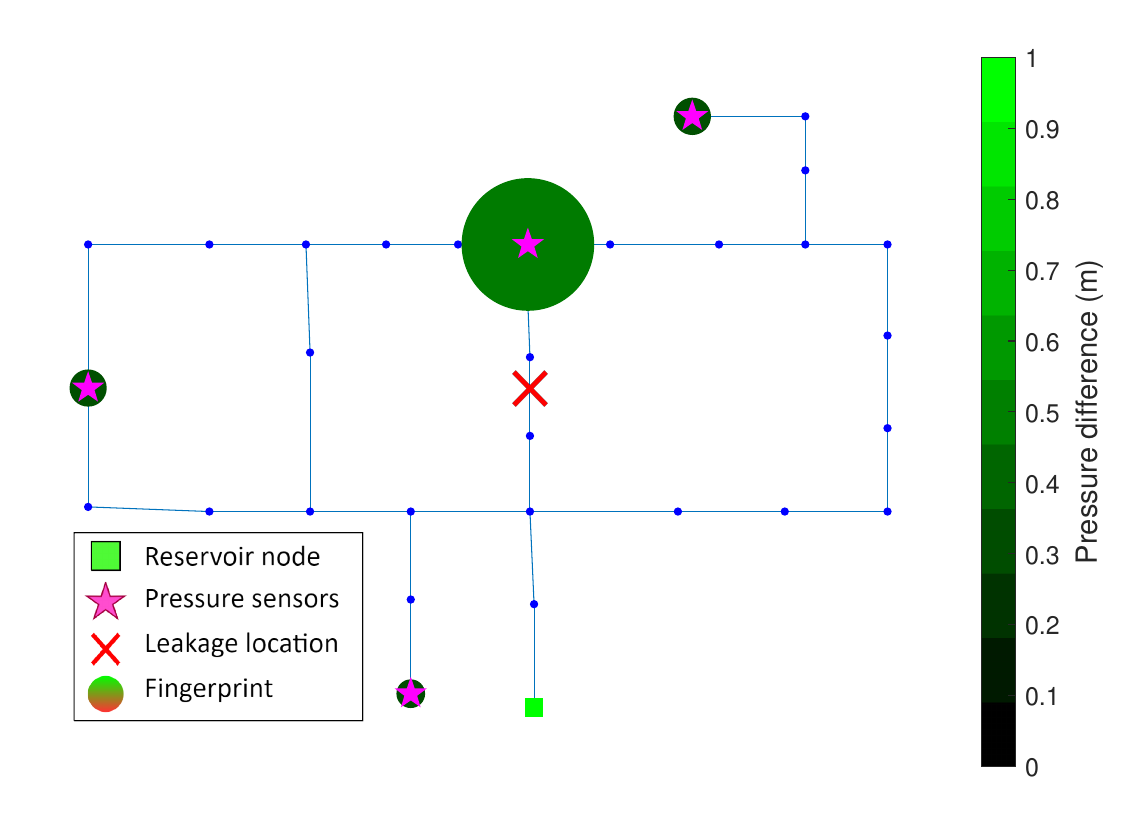}
		\caption*{(b) A leak between nodes 18 and 19.} 
	\end{minipage}
\caption{Hanoi Network: Illustration of CDFs (\refdef{def:closest_counterfactual_event_detection_fingerprint}) of sensor faults vs. leakages -- the scale corresponds to the pressure measurement change relative to the actual measurements so that a fault is not detected.}
	\label{fig:hanoi:fingerprint:examples}
\end{figure}

In the first scenario, a sensor fault (an additive offset, linearly proportional to the true value) occurs at node 13.
The Event Detection module then detects the event.
The closest counterfactual fingerprint (see Section \ref{sec:counterfactual_event_detection_fingerprint}) is then calculated and illustrated in Fig. \ref{fig:hanoi:fingerprint:examples} (a).
This indicates the change in each pressure sensor's measurement values that would result in the event not being detected.
In the case of the sensor fault, the largest change is associated with the faulty sensor. 

In the second scenario, a leakage occurs at node 17, and the Event Detection module then detects the event.
Again, the closest counterfactual fingerprint is calculated and is illustrated in Fig. \ref{fig:hanoi:fingerprint:examples} (b).
In this case, the greatest change is associated with the sensor nearest to the leakage, however, other sensor measurements should change as well, in contrast to the sensor fault case. 
\FloatBarrier

\subsection{Quantitative Performance Evaluation}
We evaluate our proposed methodology, on $204$ scenarios (three months long) with a single leak and $52$ scenarios with a single sensor fault, by first evaluating the performance of our implemented event detection in Table~\ref{table:exp:results:hanoi:eventdetection} and of the subsequent event isolation (i.e., event classification) in Table~\ref{table:exp:results:hanoi:eventclassification}. The results of the event classification baseline (see Section~\ref{sec:event_isolation:baseline}) are given in Table~\ref{table:exp:results:hanoi:eventclassification:baseline}.
Note that we only show the results for a decision tree as an event classifier, the results of a k-nearest neighbor classifier for event isolation are given in~\ref{appendix:experiments1}.

We observe a good performance of the implemented event detection method -- i.e., a very high true positive (TP) and true negative (TN) score and only very few false alarms (FP and FN). We observe only minor differences for the different types of sensor faults, but significant differences for different leak sizes -- i.e., small leakages are much harder to detect than large leakages, which is also visible in the detection delay.
Regarding the event classification, we observe a strong performance of our proposed interpretable event classifier. The baseline classifier performs significantly worse on classifying leakages but yields comparable performance (precision) on classifying sensor faults -- this suggests that identifying sensor faults is easier than identifying leakages which is reasonable since leakages affect more than one sensor in the network. The results of k-nearest neighbor as an event classifier are comparable and often a bit better. 
\begin{table}[h!]
\centering
\caption{Evaluation of the event detection method -- we report the mean an variance over all realistic Hanoi scenarios, all numbers are rounded to two decimal points.}
\small
\begin{tabular}{|c||c|c|c|c|c|}
 \hline
 \textit{Event type} & TP$\uparrow$ & FP$\downarrow$ & FN$\downarrow$ & TN$\uparrow$ & Detection Delay$\downarrow$ \\
 \hline\hline
 Leakage & $0.98 \pm 0.01$ & $0.02 \pm 0.01$ & $0.08 \pm 0.02$ & $0.92 \pm 0.02$ & $5.96 \pm 751.5$  \\
 \hline
 Small & $0.96 \pm 0.02$ & $0.04 \pm 0.02$ & $0.22 \pm 0.03$ & $0.78 \pm 0.03$ & $17.91 \pm 2063.01$ \\
 Medium & $1.0 \pm 0.0$ & $0.0 \pm 0.0$ & $0.02 \pm 0.0$ & $0.98 \pm 0.0$ & $0.13 \pm 0.59$  \\
 Large & $1.0 \pm 0.0$ & $0.0 \pm 0.0$ & $0.0 \pm 0.0$ & $1.0 \pm 0.0$ & $0.0 \pm 0.0$  \\
 \hline\hline
 Sensor Fault & $0.98 \pm 0.0$ & $0.02 \pm 0.0$ & $0.1 \pm 0.03$ & $0.9 \pm 0.03$ & $19.74 \pm 3978.62$  \\
 \hline
 Gaussian noise & $0.96 \pm 0.0$ & $0.04 \pm 0.0$ & $0.37 \pm 0.01$ & $0.63 \pm 0.01$ & $17.59 \pm 1382.76$ \\
 Power failure & $0.98 \pm 0.01$ & $0.02 \pm 0.01$ & $0.01 \pm 0.01$ & $0.99 \pm 0.01$ & $23.97 \pm 17809.53$ \\
 Offset & $1.0 \pm 0.0$ & $0.0 \pm 0.0$ & $0.0 \pm 0.0$ & $1.0 \pm 0.0$ & $0.0 \pm 0.0$ \\
 Drift & $1.0 \pm 0.0$ & $0.0 \pm 0.0$ & $0.04 \pm 0.0$ & $0.96 \pm 0.0$ & $49.14 \pm 3944.55$ \\
 \hline
\end{tabular}
\label{table:exp:results:hanoi:eventdetection}
\end{table}

\begin{table}[h!]
\centering
\caption{Evaluation of the \textit{proposed event classification method} (decision tree) -- we report the mean over the realistic Hanoi scenarios, all numbers are rounded to two decimal points.}
\small
\begin{tabular}{|c||c|c|c|}
 \hline
 \textit{Event type} & Precision$\uparrow$ & Recall$\uparrow$ & F1-Score$\uparrow$ \\
 \hline
 Leakage & $0.93$ & $0.94$ & $0.94$ \\
 \hline
 Small & $0.93$ & $0.93$ & $0.93$ \\
 Medium & $0.93$ & $0.94$ & $0.94$ \\
 Large & $0.93$ & $0.94$ & $0.93$ \\
 \hline\hline
 Sensor Fault & $0.94$ & $0.93$ & $0.94$ \\
 \hline
 Gaussian noise & $0.9$ & $0.59$ & $0.71$ \\
 Power failure & $0.94$ & $0.97$ & $0.95$ \\
 Offset & $0.94$ & $0.97$ & $0.95$ \\
 Drift & $0.94$ & $0.93$ & $0.93$ \\
 \hline
\end{tabular}
\label{table:exp:results:hanoi:eventclassification}
\end{table}
\begin{table}[h!]
\centering
\caption{Evaluation of the \textit{event classification baseline} from Section~\ref{sec:event_isolation:baseline} (decision tree) -- we report the mean over the realistic Hanoi scenarios, all numbers are rounded to two decimal points.}
\small
\begin{tabular}{|c||c|c|c|}
 \hline
 \textit{Event type} & Precision$\uparrow$ & Recall$\uparrow$ & F1-Score$\uparrow$  \\
 \hline
 Leakage & $0.55$ & $0.99$ & $0.7$ \\
 \hline
 Small & $0.55$ & $1.0$ & $0.71$ \\
 Medium & $0.55$ & $1.0$ & $0.71$ \\
 Large & $0.54$ & $0.99$ & $0.7$ \\
 \hline\hline
 Sensor Fault & $0.96$ & $0.17$ & $0.29$ \\
 \hline
 Gaussian noise & $0.95$ & $0.13$ & $0.23$ \\
 Power failure & $0.96$ & $0.18$ & $0.3$ \\
 Offset & $0.96$ & $0.17$ & $0.29$ \\
 Drift & $0.96$ & $0.17$ & $0.3$ \\
 \hline
\end{tabular}
\label{table:exp:results:hanoi:eventclassification:baseline}
\end{table}
\FloatBarrier

\section{Case Study II}\label{sec:experiments2}

\begin{figure}[h!]
	\centering
    \includegraphics[width=\linewidth]{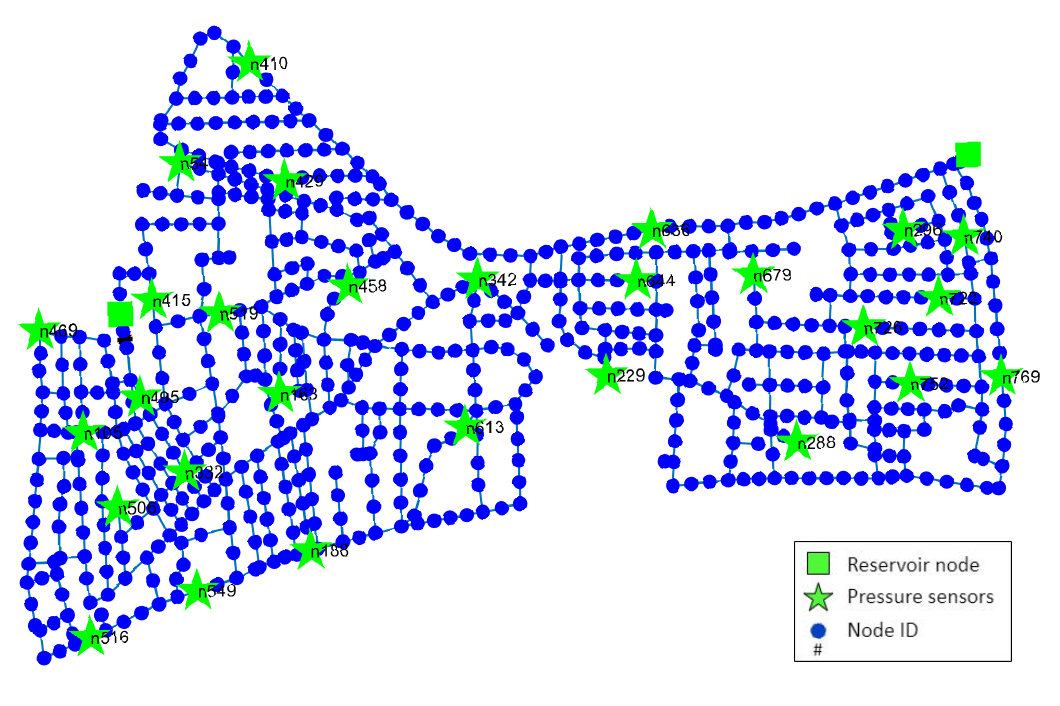} 
        \caption{The L-Town network (Area A) (Vrachimis et al., 2022) and the locations of $29$ pressure sensors.}	
 \label{fig:l_town}
\end{figure}

We consider a variant of the realistic L-Town water distribution network~\cite{Vrachimis2022} --  we demonstrate our approach on the hydraulically isolated Area A (see Fig.~\ref{fig:l_town}), by closing the valves that provide water to Area B and C. Area A consists of $661$ nodes and $766$ links. 
The benchmark assumes the existence of $29$ pressure sensors at nodes that have been optimally placed to maximize the detectability of leakages.

\subsection{Quantitative Performance Evaluation}\label{sec:experiments:data_setup}
We use two different instances of L-Town networks: One instance with artificial demand patterns -- this instance is supposed to model the modeled network by the water utility --, and an instance with (noisy) realistic demand patterns and changed hydraulic dynamics, which is achieved by adding random noise to the realistic demand patterns and closing some valves to change the dynamics -- this instance is supposed to model the real world which is a bit different from the model used by the water utility.

\begin{figure}
	\centering
	\begin{minipage}[b]{0.49\textwidth}
		\includegraphics[width=\textwidth]{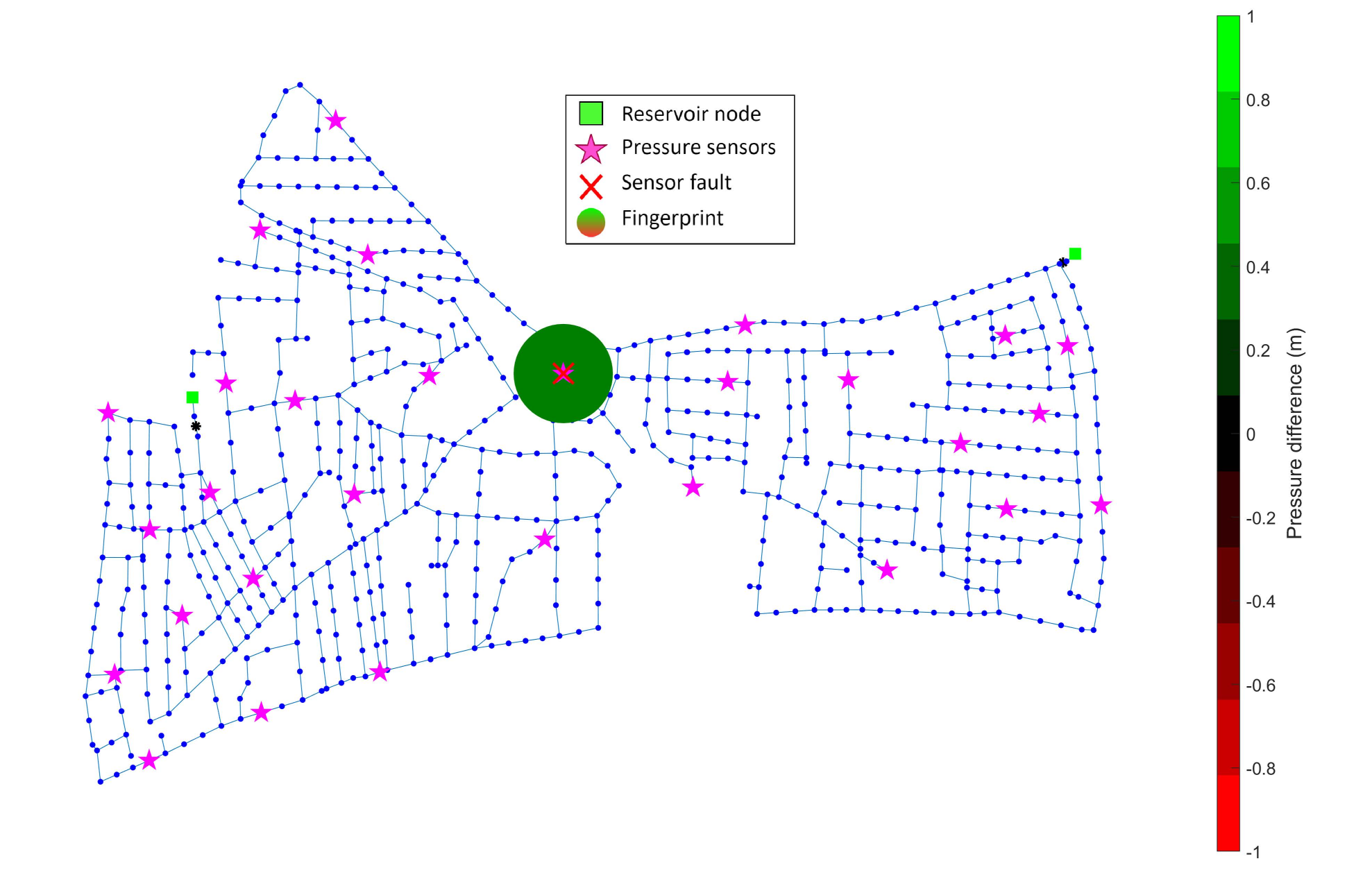}  
		\caption*{Sensor fault at sensor 'n342'.} 
	\end{minipage}
	\hfill
	\begin{minipage}[b]{0.49\textwidth}
		\includegraphics[width=\textwidth]{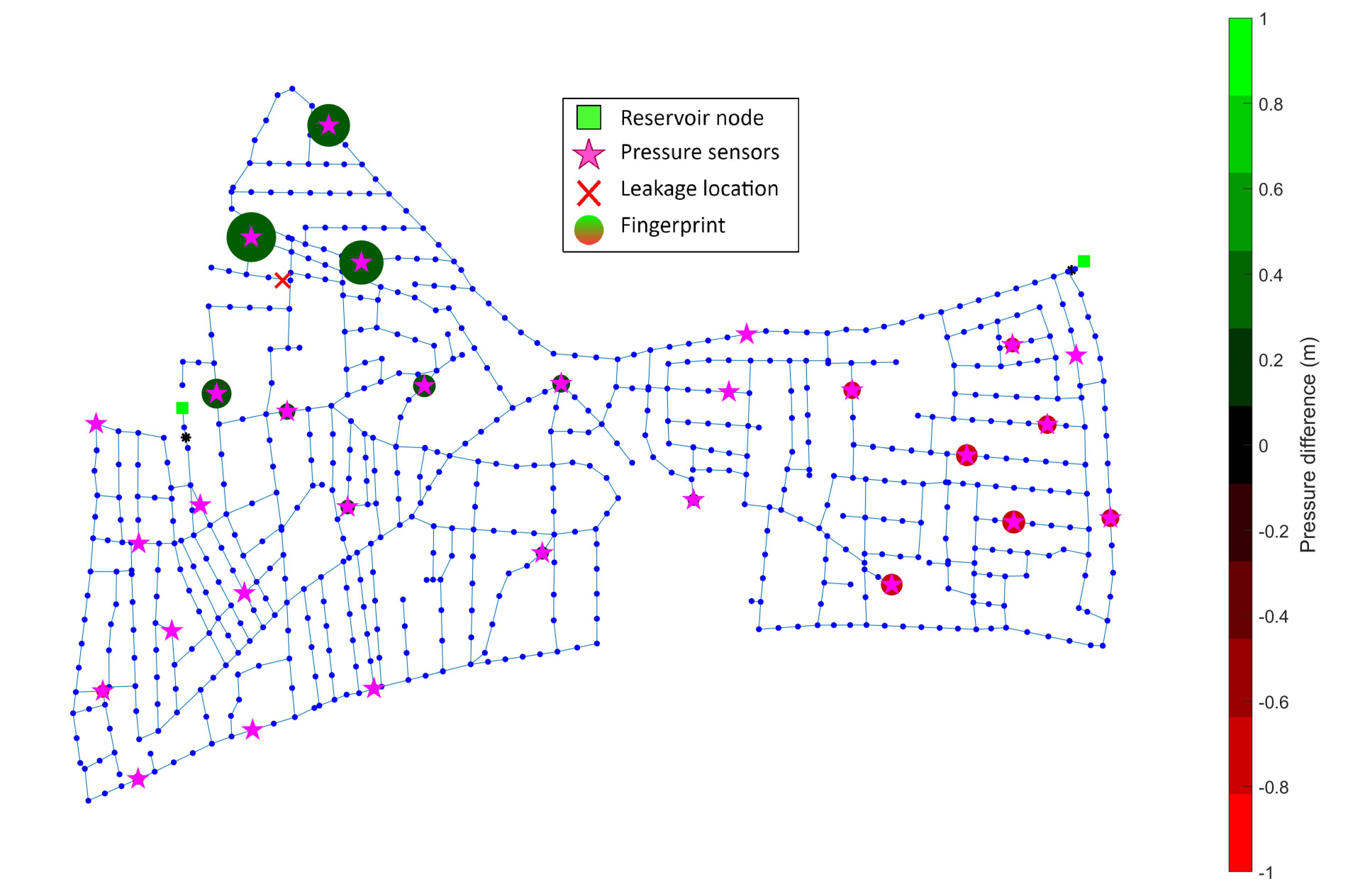}
		\caption*{Leak near node 'n66'.} 
	\end{minipage}
	\caption{L-Town network: Illustration of CDFs (\refdef{def:closest_counterfactual_event_detection_fingerprint}) of sensor faults vs. leakages -- the scale corresponds to the pressure measurement change relative to the actual measurements so that a fault is not detected.}
 	\label{fig:fingerprint:examples}
\end{figure}

We generate $377$ scenarios (each three months long) with a single sensor fault (we vary type, location, and magnitude of the fault) -- we use the same types of sensor faults as in case study 1 (see Section~\ref{sec:experiments}) -- and another $534$ scenarios with a single leakage (we vary size, location, and magnitude of the leakages, but make sure that each leakage size is equally frequent in the final collection of scenarios) -- for these, we consider the L-Town instance with artificial demand patterns.
We apply our event detection mechanism to these scenarios and compute CDFs of all raised alarms (see Fig.~\ref{fig:fingerprint:examples} for an illustrative example) -- we report metrics such as detection delay, true positives, and false alarms. 
We then use these CDFs as a labeled (recall that we know which type of anomaly was present) training data set for building the event classifier (event isolation). This event classifier itself is evaluated on new scenarios generated using the realistic instances of the L-Town network, so that we can test how well the trained model generalizes to slightly different networks: $377$ scenarios with a single sensor fault and $35$ scenarios with a single leakage -- we report precision, recall, and f1-score for both event types separately.

We evaluate the performance of our implemented event (e.g. fault) detection in Table~\ref{table:exp:results:eventdetection} -- here we only report the results on the realistic test networks.
\begin{table*}
\centering
\caption{Evaluation of the event detection method -- we report the mean and variance over the realistic L-Town scenarios, all numbers are rounded to two decimal points.}
\small
\begin{tabular}{|c||c|c|c|c|c|}
 \hline
 \textit{Event type} & TP$\uparrow$ & FP$\downarrow$ & FN$\downarrow$ & TN$\uparrow$ & Detection Delay$\downarrow$  \\
 \hline\hline
 Leakage & $0.84 \pm 0.07$ & $0.16 \pm 0.07$ & $0.28 \pm 0.01$ & $0.72 \pm 0.01$ & $1134.66 \pm 3423456.45$ \\
 \hline
 Small & $0.58 \pm 0.09$ & $0.42 \pm 0.09$ & $0.35 \pm 0.0$ & $0.65 \pm 0.0$ & $3090.18 \pm 4864772.51$ \\
 Medium & $0.94 \pm 0.01$ & $0.06 \pm 0.01$ & $0.3 \pm 0.01$ & $0.7 \pm 0.01$ & $476.75 \pm 300014.19$ \\
 Large & $1.0 \pm 0.0$ & $0.0 \pm 0.0$ & $0.2 \pm 0.02$ & $0.8 \pm 0.02$ & $0.0 \pm 0.0$ \\
 \hline\hline
 Sensor Fault & $1.0 \pm 0.0$ & $0.0 \pm 0.0$ & $0.01 \pm 0.0$ & $0.99 \pm 0.0$ & $2.34 \pm 57.38$  \\
 \hline
 Gaussian noise & $1.0 \pm 0.0$ & $0.0 \pm 0.0$ & $0.06 \pm 0.0$ & $0.94 \pm 0.0$ & $0.11 \pm 0.15$ \\
 Power failure & $1.0 \pm 0.0$ & $0.0 \pm 0.0$ & $0.0 \pm 0.0$ & $1.0 \pm 0.0$ & $0.0 \pm 0.0$ \\
 Offset & $1.0 \pm 0.0$ & $0.0 \pm 0.0$ & $0.0 \pm 0.0$ & $1.0 \pm 0.0$ & $0.0 \pm 0.0$ \\
 Drift & $1.0 \pm 0.0$ & $0.0 \pm 0.0$ & $0.0 \pm 0.0$ & $1.0 \pm 0.0$ & $10.05 \pm 171.4$ \\
 \hline
\end{tabular}
\label{table:exp:results:eventdetection}
\end{table*}

We report the results of our proposed event isolation method (i.e. event classification) in Table~\ref{table:exp:results:eventclassification}, and for a comparison the results of the baseline event classification in Table~\ref{table:exp:results:eventclassification:baseline}.
\begin{table}
\caption{Evaluation of the \textit{proposed event classification method} (decision tree) -- we report the mean over the realistic L-Town scenarios, all numbers are rounded to two decimal points.}
\centering
\small
\begin{tabular}{|c||c|c|c|}
 \hline
 \textit{Event type} & Precision$\uparrow$ & Recall$\uparrow$ & F1-Score$\uparrow$ \\
 \hline
 Leakage & $0.96$ & $0.98$ & $0.97$ \\
 \hline
 Small & $0.95$ & $0.91$ & $0.93$ \\
 Medium & $0.95$ & $0.95$ & $0.95$ \\
 Large & $0.96$ & $0.99$ & $0.97$ \\
 \hline\hline
 Sensor Fault & $0.98$ & $0.95$ & $0.97$\\
 \hline
 Gaussian noise & $0.98$ & $0.78$ & $0.87$ \\
 Power failure & $0.98$ & $1.0$ & $0.99$ \\
 Offset & $0.98$ & $1.0$ & $0.99$ \\
 Drift & $0.98$ & $1.0$ & $0.99$ \\
 \hline
\end{tabular}
\label{table:exp:results:eventclassification}
\end{table}
\begin{table}
\centering
\caption{Evaluation of the \textit{event classification baseline} from Section~\ref{sec:event_isolation:baseline} (decision tree) -- we report the mean over the realistic L-Town scenarios, all numbers are rounded to two decimal points.}
\small
\begin{tabular}{|c||c|c|c|}
 \hline
 \textit{Event type} & Precision$\uparrow$ & Recall$\uparrow$ & F1-Score$\uparrow$  \\
 \hline
 Leakage & $0.96$ & $0.89$ & $0.92$ \\
 \hline
 Small & $0.89$ & $0.32$ & $0.48$\\
 Medium & $0.96$ & $0.88$ & $0.92$ \\
 Large & $0.96$ & $0.91$ & $0.93$ \\
 \hline\hline
 Sensor Fault & $0.9$ & $0.96$ & $0.93$ \\
 \hline
 Gaussian noise & $0.89$ & $0.82$ & $0.85$ \\
 Power failure & $0.9$ & $1.0$ & $0.95$ \\
 Offset & $0.9$ & $1.0$ & $0.95$ \\
 Drift & $0.9$ & $1.0$ & $0.95$ \\
 \hline
\end{tabular}
\label{table:exp:results:eventclassification:baseline}
\end{table}
Similar to the first case study (Section~\ref{sec:experiments}), we observe a strong performance of our implemented event detection method. However, the detection delay is significantly larger and the scores for detecting small leakages are also worse compared to the first case study -- this is to be expected since the L-Town benchmark is more complex and therefore more difficult than the Hanoi benchmark. Nevertheless, detecting sensor faults seems to be easy on L-Town as well.
Regarding the event classification, we observe a strong performance of our proposed event classifier which is slightly better than the baseline classifier. Again, the baseline classifier has difficulties classifying small leakages which is reasonable since small leakages do not manifest themselves that clearly in the sensor residuals. The k-nearest neighbor classifier yields an excellent performance for all considered event types and there is no difference compared to the baseline -- this might be due to the memorization of the k-nearest neighbor classifier, which means that the classifier does not learn a general decision rule and is, therefore, less suited for practice.
Since the k-nearest neighbor classifier has a much larger computational complexity than a decision tree classifier, we recommend using the decision tree classifier.
Note that we only show the results for a decision tree as an event classifier, the results of a k-nearest neighbor classifier for event isolation are given in~\ref{appendix:experiments2}.

\section{Discussion, Conclusions and Future Work}\label{sec:conclusion}
In this work, we proposed a novel methodology and framework for interpretable event diagnosis in water distribution networks. In this context, we introduced the concept of counterfactual event fingerprints for explaining event detection and event isolation methods by means of contrasting explanations.
We also successfully demonstrated the value of our proposed methodology in a realistic use case based on the popular L-Town benchmark. 

Based on this work, there exist a couple of potential directions for future research:
\begin{itemize}
    \item Our proposed methodology for interpretable event isolation is limited to known types of events -- i.e. it can not deal with new types of events that were not available at training time. Since we can not foresee all possible events that might occur in reality, it is of interest to extend our proposed methodology to support unknown events as well -- e.g. by adding an explainable reject option to the event isolation method.
    \item An essential next step in this research involves validating its practical viability. A user-centric evaluation will be instrumental in demonstrating the methodology's real-world applicability, bridging the gap between theory and the operational needs of our intended users in real-world scenarios.
    \item Although our implemented methodology is purely data-driven, it shows a strong performance. However, it is not guaranteed that the results will behave according to the hydraulic laws -- e.g. negative pressure values might be predicted. In order to make the methodology consider the hydraulic laws and principles, physics-informed machine learning could be explored.
    \item In this work, we implemented the \textit{counterfactual event detection fingerprint} by considering each time step separately -- i.e. for each time step either an alarm was raised or not. Consequently, the computed fingerprints are valid for a particular time point only -- an alternative would be to integrate over time and compute a fingerprint for a time interval instead of a single time point, which would reduce the amount of information that has to be handled by the operator and therefore improve the systems' usability.
\end{itemize}

\section*{Acknowledgment}
This research was supported by the Ministry of Culture and Science NRW (Germany) as part of the Lamarr Fellow Network. This publication reflects the views of the authors only.
We also gratefully acknowledge funding from the European Research Council (ERC) under the ERC Synergy Grant Water-Futures (Grant agreement No. 951424), and funding by the German federal state of North Rhine-Westphalia for the research training group ``Dataninja'' (Trustworthy AI for Seamless Problem Solving: Next Generation Intelligence Joins Robust Data Analysis).

\appendix
\FloatBarrier
\section{Case Study I}\label{appendix:experiments1}
\begin{table}[h!]
\centering
\caption{Evaluation of the \textit{proposed event classification method} (k-nearest neighbor classifier) -- we report the mean over the realistic Hanoi scenarios, all numbers are rounded to two decimal points.}
\small
\begin{tabular}{|c||c|c|c|}
 \hline
 \textit{Event type} & Precision$\uparrow$ & Recall$\uparrow$ & F1-Score$\uparrow$  \\
 \hline
 Leakage & $0.99$ & $0.99$ & $0.99$ \\
 \hline
 Small & $0.99$ & $0.99$ & $0.99$ \\
 Medium & $0.99$ & $0.99$ & $0.99$ \\
 Large & $0.99$ & $0.99$ & $0.99$ \\
 \hline\hline
 Sensor Fault & $0.99$ & $0.99$ & $0.99$ \\
 \hline
 Gaussian noise & $0.99$ & $0.95$ & $0.97$ \\
 Power failure & $0.99$ & $1.0$ & $0.99$ \\
 Offset & $0.99$ & $1.0$ & $0.99$ \\
 Drift & $0.99$ & $1.0$ & $0.99$ \\
 \hline
\end{tabular}
\label{table:exp:results:hanoi:eventclassification2}
\end{table}

\begin{table}
\centering
\caption{Evaluation of the \textit{event classification baseline} from Section~\ref{sec:event_isolation:baseline} (k-nearest neighbor classifier) -- we report the mean over the realistic Hanoi scenarios, all numbers are rounded to two decimal points.}
\small
\begin{tabular}{|c||c|c|c|}
 \hline
 \textit{Event type} & Precision$\uparrow$ & Recall$\uparrow$ & F1-Score$\uparrow$ \\
 \hline
 Leakage & $0.73$ & $0.72$ & $0.72$ \\
 \hline
 Small & $0.73$ & $0.7$ & $0.72$ \\
 Medium & $0.73$ & $0.72$ & $0.73$ \\
 Large & $0.73$ & $0.71$ & $0.72$ \\
 \hline\hline
 Sensor Fault & $0.72$ & $0.74$ & $0.73$\\
 \hline
 Gaussian noise & $0.75$ & $0.85$ & $0.79$ \\
 Power failure & $0.72$ & $0.72$ & $0.72$ \\
 Offset & $0.72$ & $0.73$ & $0.72$ \\
 Drift & $0.72$ & $0.74$ & $0.73$ \\
 \hline
\end{tabular}
\label{table:exp:results:hanoi:eventclassification2:baseline}
\end{table}
\FloatBarrier

\section{Case Study 2}\label{appendix:experiments2}
\begin{table}[h!]
\centering
\caption{Evaluation of the \textit{proposed event classification method} (k-nearest neighbor classifier) -- we report the mean over the realistic L-Town scenarios, all numbers are rounded to two decimal points.}
\small
\begin{tabular}{|c||c|c|c|}
 \hline
 \textit{Event type} & Precision$\uparrow$ & Recall$\uparrow$ & F1-Score$\uparrow$ \\
 \hline
 Leakage & $1.0$ & $1.0$ & $1.0$ \\
 \hline
 Small & $1.0$ & $0.99$ & $0.99$ \\
 Medium & $1.0$ & $1.0$ & $1.0$ \\
 Large & $1.0$ & $1.0$ & $1.0$ \\
 \hline\hline
 Sensor Fault & $1.0$ & $1.0$ & $1.0$\\
 \hline
 Gaussian noise & $1.0$ & $0.99$ & $1.0$ \\
 Power failure & $1.0$ & $1.0$ & $1.0$ \\
 Offset & $1.0$ & $1.0$ & $1.0$ \\
 Drift & $1.0$ & $1.0$ & $1.0$ \\
 \hline
\end{tabular}
\label{table:exp:results:ltown:eventclassification2}
\end{table}
\begin{table}[h!]
\centering
\caption{Evaluation of the \textit{event classification baseline} from Section~\ref{sec:event_isolation:baseline} (k-nearest neighbor classifier) -- we report the mean over the realistic L-Town scenarios, all numbers are rounded to two decimal points.}
\small
\begin{tabular}{|c||c|c|c|}
 \hline
 \textit{Event type} & Precision$\uparrow$ & Recall$\uparrow$ & F1-Score$\uparrow$  \\
 \hline
 Leakage &$1.0$ & $1.0$ & $1.0$ \\
 \hline
 Small & $1.0$ & $0.9$ & $0.95$ \\
 Medium & $1.0$ & $0.99$ & $1.0$ \\
 Large & $1.0$ & $1.0$ & $1.0$ \\
 \hline\hline
 Sensor Fault & $1.0$ & $1.0$ & $1.0$\\
 \hline
 Gaussian noise & $1.0$ & $0.99$ & $1.0$ \\
 Power failure & $1.0$ & $1.0$ & $1.0$ \\
 Offset & $1.0$ & $1.0$ & $1.0$ \\
 Drift & $1.0$ & $1.0$ & $1.0$ \\
 \hline
\end{tabular}
\label{table:exp:results:ltown:eventclassification2:baseline}
\end{table}
\FloatBarrier


\bibliographystyle{elsarticle-num}
\bibliography{references,references_hydraulics}


\end{document}